\documentclass[%
reprint,
 amsmath,amssymb,
 aps,
 prl,
 placeins,
]{revtex4-2}

\usepackage{graphicx}
\usepackage{dcolumn}
\usepackage{bm}
\usepackage{float}
\usepackage{subcaption}
\usepackage{multirow}
\usepackage{comment}
\usepackage{dsfont}

\def\ie{{\frenchspacing\it i.e.}}
\def\eg{{\frenchspacing\it e.g.}}
\def\etc{{\frenchspacing\it etc.}}

\newcommand{\mat}[1]{\mathbf{#1}}
\def\dim{{\rm dim}}
\def\e{\textbf{e}}
\def\th{{\boldsymbol{\theta}}}

\def\neff{n_{\rm eff}}

\def\I{{\bf I}}
\def\M{{\cal M}}
\def\n{\textbf{n}}

\def\shat{{\hat{s}}}
\def\p{\textbf{p}}
\def\q{\textbf{q}}
\def\x{\textbf{x}}
\def\y{\textbf{y}}
\def\T{{\cal S}}
\def\Tbar{\bar{\T}}
\def\L{L}

\def\spose#1{\hbox to 0pt{#1\hss}}
\def\simlt{\mathrel{\spose{\lower 3pt\hbox{$\mathchar"218$}}
     \raise 2.0pt\hbox{$\mathchar"13C$}}}
\def\simgt{\mathrel{\spose{\lower 3pt\hbox{$\mathchar"218$}}
     \raise 2.0pt\hbox{$\mathchar"13E$}}}
\def\simpropto{\mathrel{\spose{\lower 3pt\hbox{$\mathchar"218$}}
     \raise 2.0pt\hbox{$\propto$}}}

\def\beq#1{\begin{equation}\label{#1}}
\def\eeq{\end{equation}}
\def\beqa#1{\begin{eqnarray}\label{#1}}
\def\eeqa{\end{eqnarray}}
\def\eq#1{equation~(\ref{#1})}

\def\fig#1{FIG.~\ref{#1}}
\def\Fig#1{FIG.~\ref{#1}}

\begin{document}

\preprint{APS/123-QED}

\title{AI Poincar\'{e}: Machine Learning Conservation Laws from Trajectories}

\author{Ziming Liu}
\author{Max Tegmark}%
\affiliation{%
 Department of Physics, Massachusetts Institute of Technology, Cambridge, USA
}%

\date{\today}

\begin{abstract}
We present {\it AI Poincar\'{e}}, a machine learning algorithm for auto-discovering conserved quantities using trajectory data from unknown dynamical systems.
We test it on five Hamiltonian systems, including the gravitational 3-body problem, and find that it discovers not only all exactly conserved quantities, but also periodic orbits, phase transitions and breakdown timescales for approximate conservation laws.
\end{abstract}
\maketitle

\section{Introduction}
\vskip-4mm

While machine learning has contributed to many physics advances, 
such as improving the speed or quality of numerical simulations, laboratory experiments and astronomical observations~\cite{van_Nieuwenburg_2017,Hezaveh_2017,Sun_2018,baldi2014searching,pang2018equation,RAISSI2019686,albergo2019flowbased}, 
a more ambitious goal is to design intelligent machines to make new scientific discoveries such as physical symmetries ~\cite{decelle_localsymmetry,mattheakis2019physical,bondesan2019learning,mototake2019interpretable,PhysRevResearch.2.033499}
and formulas via symbolic regression~\cite{kim2019integration,lu2019extracting,cranmer2020discovering,Udrescueaay2631,udrescu2020ai}. 
In this spirit, the goal of the present paper is to auto-discover conservation laws from trajectories of dynamical systems.

Physicists have traditionally derived conservation laws in a {\it model-driven} way, such as when Poincar\'{e} 
proved \cite{poincare} that the 3D gravitational 3-body problem has only 10 conserved quantities.
In contrast, this paper aims to discover conservation laws in a {\it data-driven} way, 
using only observed trajectory data as input while treating the underlying dynamical equations as unknown. 

To the best of our knowledge,~\cite{mototake2019interpretable,PhysRevResearch.2.033499} have pursued the goal closest to ours, but with an orthogonal approach detecting symmetry with an auto-encoder and Siamese neural networks respectively, requiring hand-crafted features precluding full automation, and testing on relatively simple examples. Other work linking conservation laws and machine learning ~\cite{Cranmer2020LagrangianNN,Greydanus2019HamiltonianNN,lutter2018deep,mattheakis2019physical} focus on embedding physical inductive biases (such as the existence of a Hamiltonian or Lagrangian) into machine learning, but not the other way around to apply machine learning for auto-discovery of conservation laws. 

Our ambitious goal of automating  conservation law discovery is enabled by recent 
machine-learning progress~\cite{Saremi2019NeuralEB} for sampling {\it manifolds}, which are intimately related to dynamical systems as
summarized in Table~\ref{tab:dynamical_manifold}:
viewing each state as a point in a phase space $\mathbb{R}^N$,
the topological closure of the set of all states on a trajectory form a manifold $\M\subset \mathbb{R}^N$. Each conservation law removes one degree of freedom from the dynamical system and one dimension from $\M$, so the number of conserved quantities is simply $N$ minus the dimensionality of $\M$ \cite{goldstein:mechanics}.
The local tangent space of $\M$ represents all local displacements allowed by conservation laws, while the space perpendicular to the tangent space is spanned by gradients of conserved quantities. 

\begin{table}[ht]
    \centering
    \caption{Manifold/dynamical system correspondence}
     \vskip-3mm
    \begin{tabular}{|l|l|}\hline
     Manifold    &  Dynamical System \\\hline
      Dimensionality reduction   & Conservation law \\
      Tangent space &Conserved quantity isosurface  \\
      Orthogonal space & Gradients of conserved quantities\\\hline
    \end{tabular}
    \label{tab:dynamical_manifold}
    \vskip-5mm
\end{table}

We introduce our notation and AI Poincar\'{e} algorithm in the Methods section. In the Results section, we  apply AI Poincar\'{e} to five Hamiltonian systems to test its ability to discover conserved quantities (numerically and symbolically), periodic orbits, phase transitions, and conservation breakdown timescales.

\begin{figure}[htbp]
\vskip-3mm
\includegraphics[width=1\linewidth]{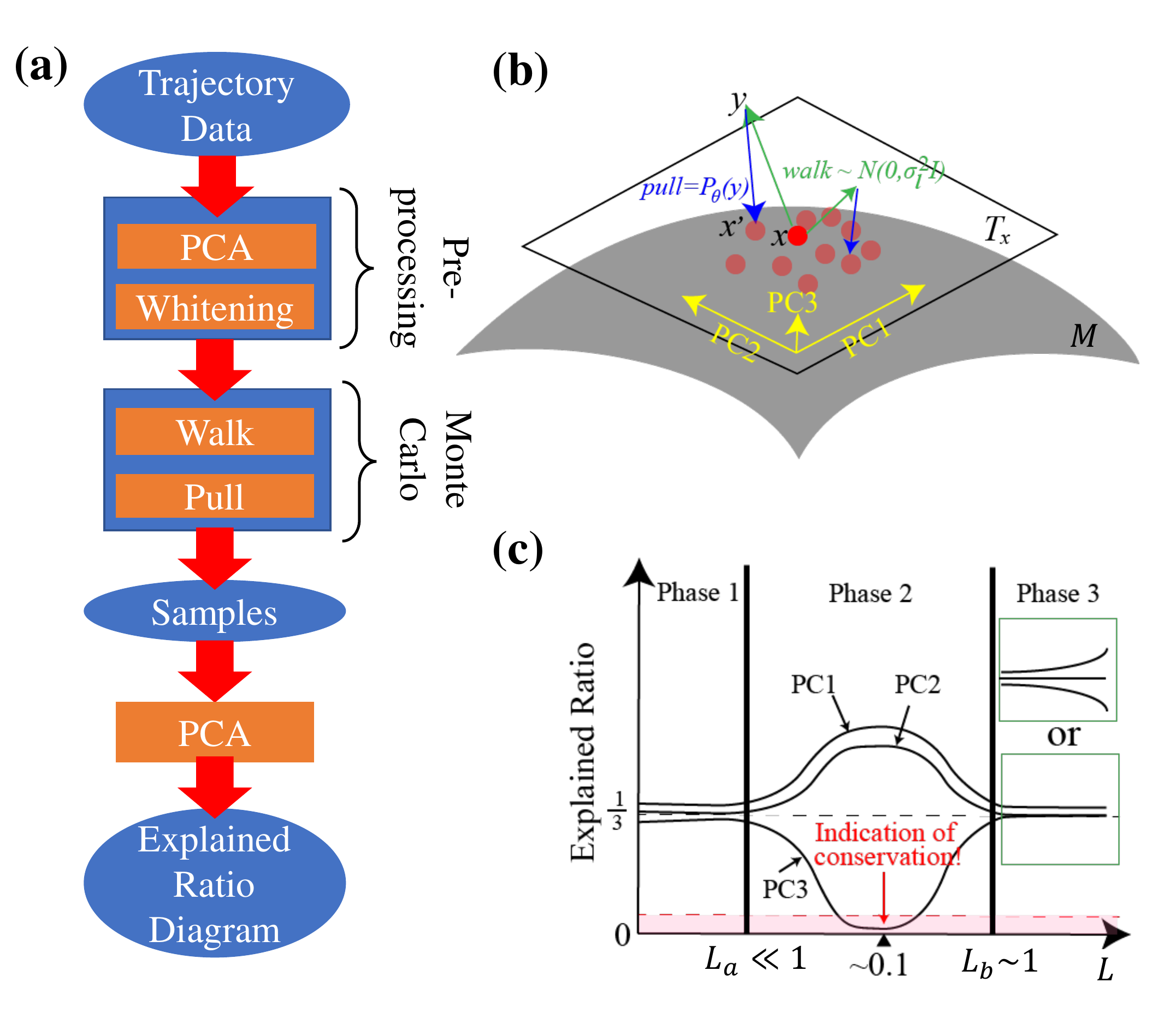}
\vskip-6mm
\caption{The AI Poincar\'{e} algorithm: (a) overall workflow, (b) walk-pull Monte Carlo module, (c) typical explained ratio diagram, with Phase 2 revealing conserved quantities. 
}
\vskip-5mm
\label{fig:ai_poincare}
\end{figure}
\vskip-30mm\section{Method}
\label{sec:method}
\vskip-3mm

{\bf Problem and notation:}
Consider a dynamical system whose state vector $\x\in\mathbb{R}^N$ evolves according to an ordinary differential equation ${d\x\over dt}=f(\x,t)$ for some smooth function $f$.
In physics, dynamical systems can often be written with $\mat{x}$ as the
concatenation of vectors of generalized coordinates $\mat{q}$ and momenta $\mat{p}$ and $N$ is even.
For the special case of {\it Hamiltonian systems}, important in classical mechanics~\cite{goldstein:mechanics}, there exists a Hamiltonian function 
$H_0(\q,\p)$ such that 
\begin{equation}
    \frac{d\mat{p}^{(i)}}{dt} =- \frac{\partial H_0}{\partial \mat{q}^{(i)}},\ \frac{d\mat{q}^{(i)}}{dt}=\frac{\partial H_0}{\partial \mat{p}^{(i)}}\quad (i=1,...,k).
\end{equation}
Conservation laws are important in physics, common examples including conservation of energy ($H_0$), momentum, angular momentum, Runge-Lenz vector, magnetic moment, {\etc}
We express a set of independent conservation laws as 
$H_j[\x(t)]=h_j$, 
$j=0,...,n-1$, valid 
exactly or approximately. 
Each conservation law can be understood as a mathematical constraint that slices the original $n$-dimensional phase space into a family of iso-surfaces. We define the \textit{permissible state manifold} (PSM) as 
$\M\equiv\{\x\in\mathbb{R}^N|H_j(\x)=h_j\}$, 
\ie, as the set of states allowed by all conservation laws. It is clear that $\dim(\M)=N-n$~\cite{goldstein:mechanics}, since each conservation law removes one degree of freedom from the system. In practice, however, only trajectory data rather than the full PSM is observed, which motivates us to define the \textit{trajectory set} as $\T=\{\mat{x}(t)|t\geq 0\}$. 
If the system is ergodic, $\T$ will be everywhere dense in $\M$ so that $\M$ is the closure $\bar{\T}$ of $\T$,
implying the identity $n=N-\dim(\Tbar)$. 
This generalizes even beyond traditional physics contexts: for example, if $\x$  
contains the pixel colors in a grey-scale video, then the color of a  pixel that always remains black is a conserved quantity.

{\bf AI Poincar\'{e}} 
We present a machine learning algorithm, \textit{AI Poincar\'{e}}, that uses $S$ to compute an estimator $\shat$ of the dimensionality $s\equiv\dim(\Tbar)$, thus obtaining an estimator $\neff\equiv N-\shat$ of $n$.
If $\neff>0$, it suggests the existence of $\neff$ hitherto undiscovered conservation laws.
If $\neff=0$, the system is not Hamiltonian since it lacks even a single conserved quantity $H_0$.

Although manifold learning is an active field
and has developed powerful tools to explore and visualize the latent structure of low-dimensional submanifolds of high-dimensional spaces~\cite{Tenenbaum2319,Roweis2323,Maaten2008VisualizingDU,LUO20032213,pca_review,ae},
they either focus on performance on downstream tasks (\eg, image/video generation) where the dimensionality is a user-specified input parameter, or perform not as well as the proposed method here on our task. The supplemental material compares the performance 
with the PCA, auto-encoder and fractal methods.
If we had perfect noiseless samples forming a dense set $\T$ in $\M$, then
we could simply determine the manifold dimensionality $s$ as the rank (number of nonzero eigenvalues) of 
the covariance matrix of the samples in an infinitesimal environment of a random sample, where the manifold can be approximated by a hyperplane. In practice, we cannot probe infinitesimal scales because we have only a finite number of points, yet we must avoid large scales where manifold curvature is important; our method handles these complications by 
 treating dimensionality $s$ as a renormalized quantity that is a function of length scale $\L$.

AI Poincar\'{e} consists of three modules as illustrated in \fig{fig:ai_poincare} (a):
(1) Pre-processing (prewhitening and optional dimensionality reduction), 
(2) local Monte Carlo sampling of $\M$ and 
(3) linear dimensionality estimation from these samples using PCA explained ratios. 
The prewhitening performs an affine transformation such that the points in $\T$ have zero mean and covariance matrix $\I$, the identity matrix. 
If any eigenvalues of the original covariance matrix vanish, then the corresponding eigenvectors $\e_i$ define linear conserved
quantities $H_i(\x)=\e_i\cdot\x$, and we remove these dimensions before proceeding. The supplementary material gives further technical details of the preprocessing module.

Our module for Monte Carlo sampling the manifold benefits from the neural empirical Bayes technique in the machine learning literature~\cite{Saremi2019NeuralEB}. It consists of two steps, as illustrated in \fig{fig:ai_poincare}  (b): 
\begin{itemize}
\itemsep0mm
    \item Walk step $\x\mapsto\y$: perturb a state vector $\mat{x}\in\T$  by adding isotropic Gaussian noise $\n$
    with zero mean and covariance $L\,\I$.
    \item Pull step $\y\mapsto\x'$: pull the ``noisy" state $\mat{y}$ back toward the manifold.
    We do this by training the parameters $\th$ of a feedforward neural network 
    implementing a  ``pull function" $P_\th$ mapping $\mathbb{R}^N$ to $\mathbb{R}^N$, 
    minimizing the loss function
    \begin{equation}\label{eq:loss}
        Loss(\th)=\frac{1}{N_{s}}\sum_{i=1}^{N_{s}} |P_\th(\y_i)-\x_i|_2^2,
    \end{equation}
Where $N_s$ is the number of samples. Since the best the neural network can do is learn to orthogonally project 
points back onto the manifold, the pulled-back points
$\x'_i\equiv P_\th(\y_i)$
characterize the local tangent space when $\L$ is appropriately chosen
~\cite{Saremi2019NeuralEB}.
\end{itemize}
The Supplementary Material provides further intuition and illustrations regarding how the preprocessing walk/pull steps work.

{\bf Explained Ratio Diagram}
The output of the AI Poincar\'{e} algorithm is the {\it explained ratio diagram} (ERD), 
showing the fraction of variance explained by each principal component (its eigenvalue divided by the eigenvalue sum) as a function of noise length scale 
$\L$, revealing the dimensionality of $\cal M$ and hence
the number of conserved quantities.
A typical ERD is shown in \fig{fig:ai_poincare} (c), revealing three phases separated by phase transitions at $L_a\ll 1$ and $L_b\sim 1$. 
In the intermediate phase $L_a\simlt\L\simlt L_b$, some principal component(s) drop near zero and are identified as sub-manifold structure (conservation laws). 
In the other two phases, all explained ratios are $\sim 1/n$ because the Monte Carlo sample covariance matrix $\sim\I$: 
on large scales because almost the whole (prewhitened) manifold is sampled and on small scales because of roughly isotropic noise. 

\begin{figure*}[ht]
	\centerline{\includegraphics[width=1\linewidth]{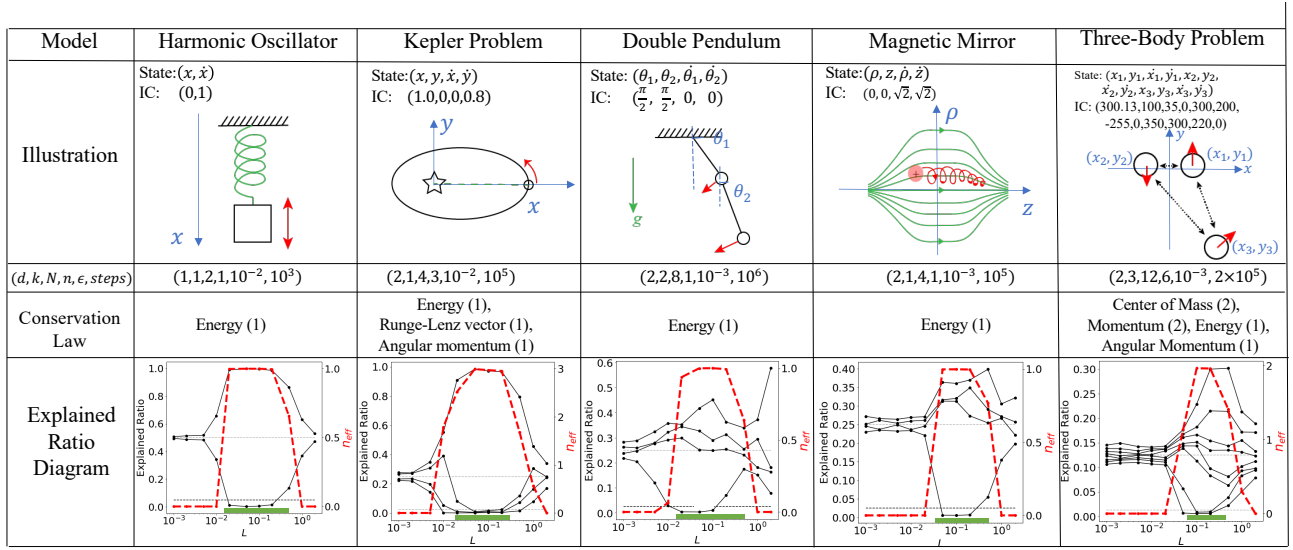}}
	\vskip-4.8cm
	\caption{Five Hamiltonian systems used to test AI Poincar\'{e}: Harmonic oscillator, Kepler problem, double pendulum, magnetic mirror and three-body problem. $d$ is the dimensionality of the Euclidean space, $k$ the number of bodies, $N\equiv 2kd$ is the phase space dimensionality, and $n$ the ground truth number of conservation laws. In the bottom panel, the red dashed curve shows the effective number of conserved quantities $\neff$ defined by \protect\eq{neffEq}, and the green region on the $L$-axis shows $L$-range that gets $n_{eff}$ correct (after rounding to the nearest integer). The horizontal dashed lines shows $1/N$ and $0.1/N$; any principal component explaining less than a fraction $0.1/N$ of the total variance at any $\L$ is considered evidence for a conservation law.}
	\label{fig:hamiltonian}
\end{figure*}

\begin{table*}[htbp]
	\centering
	\begin{tabular}{|c|c|c|}\hline
		System& Conserved quantity&Formula found?\\\hline
		Harmonic Oscillator & $H=\frac{1}{2}(x^2+{\dot x}^2)$ & Yes\\\hline
		\multirow{3}{*}{Kepler Problem}& $H=\frac{1}{2}({\dot x}^2+{\dot y}^2)-\frac{1}{\sqrt{x^2+y^2}}$ & Yes\\\cline{2-3}
		& $L_{angular}=x{\dot y}-y{\dot x}$ & Yes\\\cline{2-3}
		& $A={\rm arg}(L(-{\dot y},{\dot x})-\hat{r})$ & No\\\hline
		\multirow{2}{*}{Double Pendulum} & (Small angle) $H_s=10\theta_1^2+5\theta_2^2+{\dot\theta_1}^2+\frac{1}{2}{\dot\theta_2}^2+{\dot\theta_1}{\dot\theta_2}$ & Yes\\\cline{2-3}
		& (Large angle) $H_l=-20{\rm cos}\theta_1-10{\rm cos}\theta_2+{\dot\theta_1}^2+\frac{1}{2}{\dot \theta_2}^2+{\dot\theta_1}{\dot\theta_2}{\rm cos}(\theta_1-\theta_2)$ & No\\\hline
		Magnetic Mirror & $H=\frac{1}{2}({\dot\rho}^2+{\dot z}^2)+\frac{1}{2}(\rho^2+\frac{1}{5}z^2+\rho^2z^2)$ & Yes\\\hline
		\multirow{6}{*}{Three Body Problem}
		& $H=\sum_{i=1}^3 \frac{m}{2}({\dot x}_i^2+{\dot y}_i^2)-m^2(\frac{1}{r_{12}}+\frac{1}{r_{13}}+\frac{1}{r_{23}}),\quad m={\rm 5\times 10^6}$ & No\\\cline{2-3}
		&$x_c=\frac{1}{3}(x_1+x_2+x_3)$ & Yes\\\cline{2-3}
		& $y_c=\frac{1}{3}(y_1+y_2+y_3)$ & Yes\\\cline{2-3}
		& ${\dot x}_c=\frac{1}{3}({\dot x}_1+{\dot x}_2+{\dot x}_3)$ & Yes\\\cline{2-3}
		& ${\dot y}_c=\frac{1}{3}({\dot y}_1+{\dot y}_2+{\dot y}_3)$ & Yes\\\cline{2-3}
		& $L_{angular}=\sum_{i=1}^3 x_i{\dot y}_i-y_i{\dot x}_i$ & Yes\\\hline
	\end{tabular}
	\caption{Symbolic formulas for 10 of the 13 conservation laws were discovered using AI Feynman.}
	\label{tab:symbolic}
\end{table*}

\clearpage
\section{Results}\label{sec:results}

{\bf Numerical experiments}
We test AI Poincar\'{e} on trajectories from five well-studied Hamiltonian systems: the 1D harmonic oscillator, the 2D Kepler problem, the double pendulum, the 2D magnetic mirror and the 2D three-body problem, as defined in Table~\ref{tab:symbolic} and illustrated in \fig{fig:hamiltonian}. 
We compute trajectories for the five systems with a the 4th-order Runge-Kutta integrator 
at $N_{step}=$\{$10^3$,$10^5$,$10^6$,$10^5$,$2\times$$10^5$\} timesteps of size $\epsilon=$\{$10^{-2}$,$10^{-2}$,$10^{-3}$,$10^{-3}$,$10^{-3}$\}, using the initial conditions listed in \fig{fig:hamiltonian}.
We parametrize the pull function $P_\th$ as a feedforward neural network with two 
hidden layers containing 256 neurons each, and train it for each $\L$ using the ADAM optimizer~\cite{Kingma2015AdamAM}  with learning rate $10^{-3}$ for 5,000 steps.
We repeat the walk+pull Monte Carlo process jumping $10^4$ times from the trajectory midpoint.

{\bf Basic results}
The resulting explained ratios (\fig{fig:hamiltonian}, bottom) 
show a consistent valley around $\L=0.1$ revealing the number of conserved quantities. 
The number of conservation laws discovered by AI Poincar\'{e} is seen to agree with the ground truth (TABLE~\ref{tab:symbolic}) for all five systems if we simply define the criterion for conservation law discovery as an explained ratio that is an order of magnitude below baseline ($0.1/N$; dashed black line in the figure).
For the three-body problem, the first four conservation laws are linear and hence discovered already in our preprocessing step. As shown in the supplemental material, these results are robust to changing the starting point for the walk+pull process, and outperform the PCA, autoencoder and fractal methods for dimensionality estimation. 

{\bf Symbolic formula discovery} TABLE~\ref{tab:symbolic} (right column) shows that we can not only auto-discover 
that a conservation law exists, but in many cases also a symbolic formula for it:
we did this by applying the AI Feynman symbolic regression algorithm 
\cite{Udrescueaay2631,udrescu2020ai} to our trajectory data.
Since any function of a conserved quantity is also conserved, 
we require some form of ``gauge fixing" to make the symbolic regression problem well posed.
Here we simply require states on two chosen trajectories to have conserved value 1 and 2; this approach can undoubtedly be greatly improved, \eg, by requiring that gradient directions match those of our pull function.

\begin{figure}[t]
    \centering
    \begin{subfigure}[b]{0.23\textwidth}
            \centering \includegraphics[width=\textwidth]{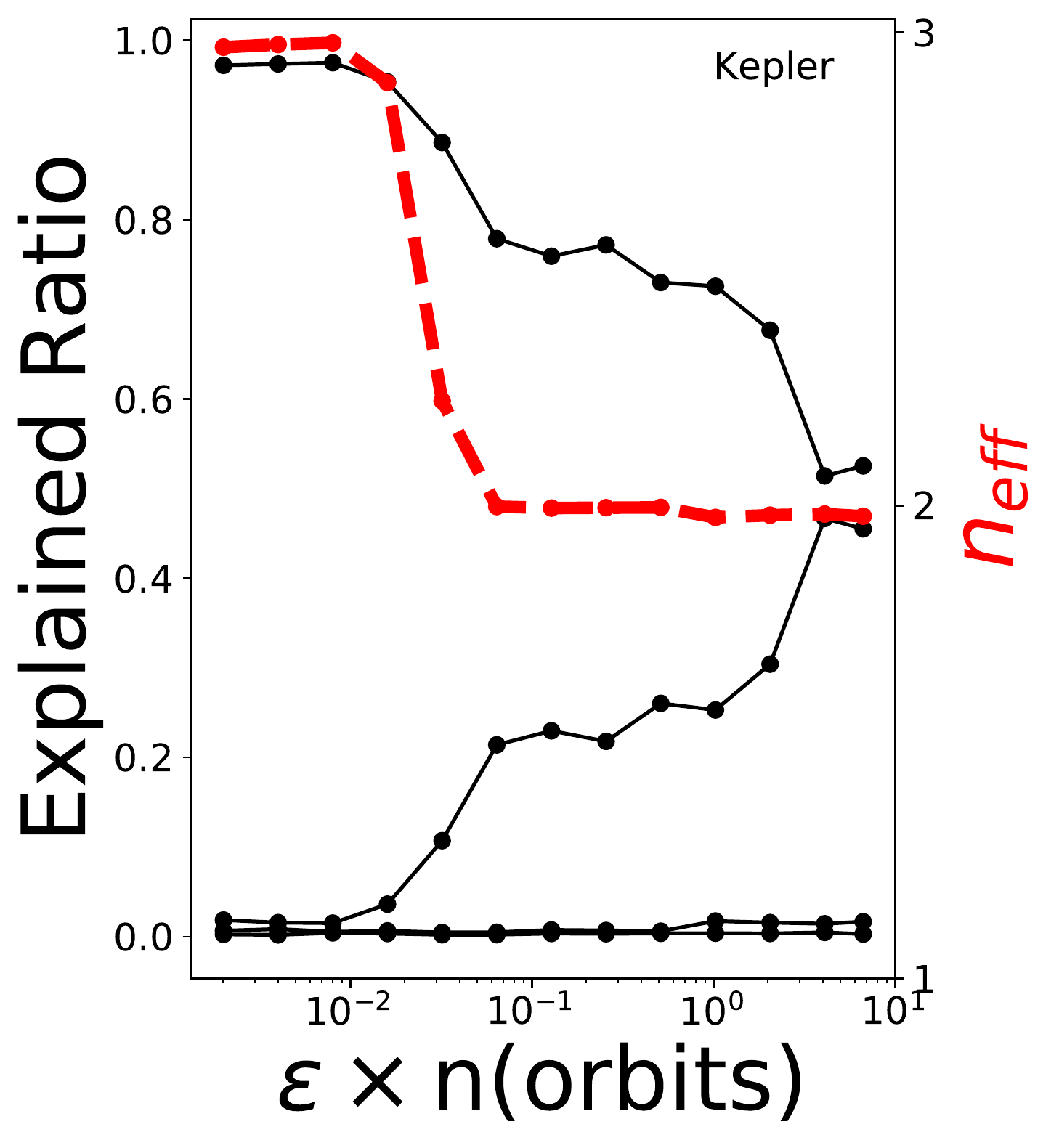}
  	    \vskip-2mm
            \caption{Kepler Problem}
    \end{subfigure}
    \hfill    
    \begin{subfigure}[b]{0.23\textwidth}  
            \centering  \includegraphics[width=\textwidth]{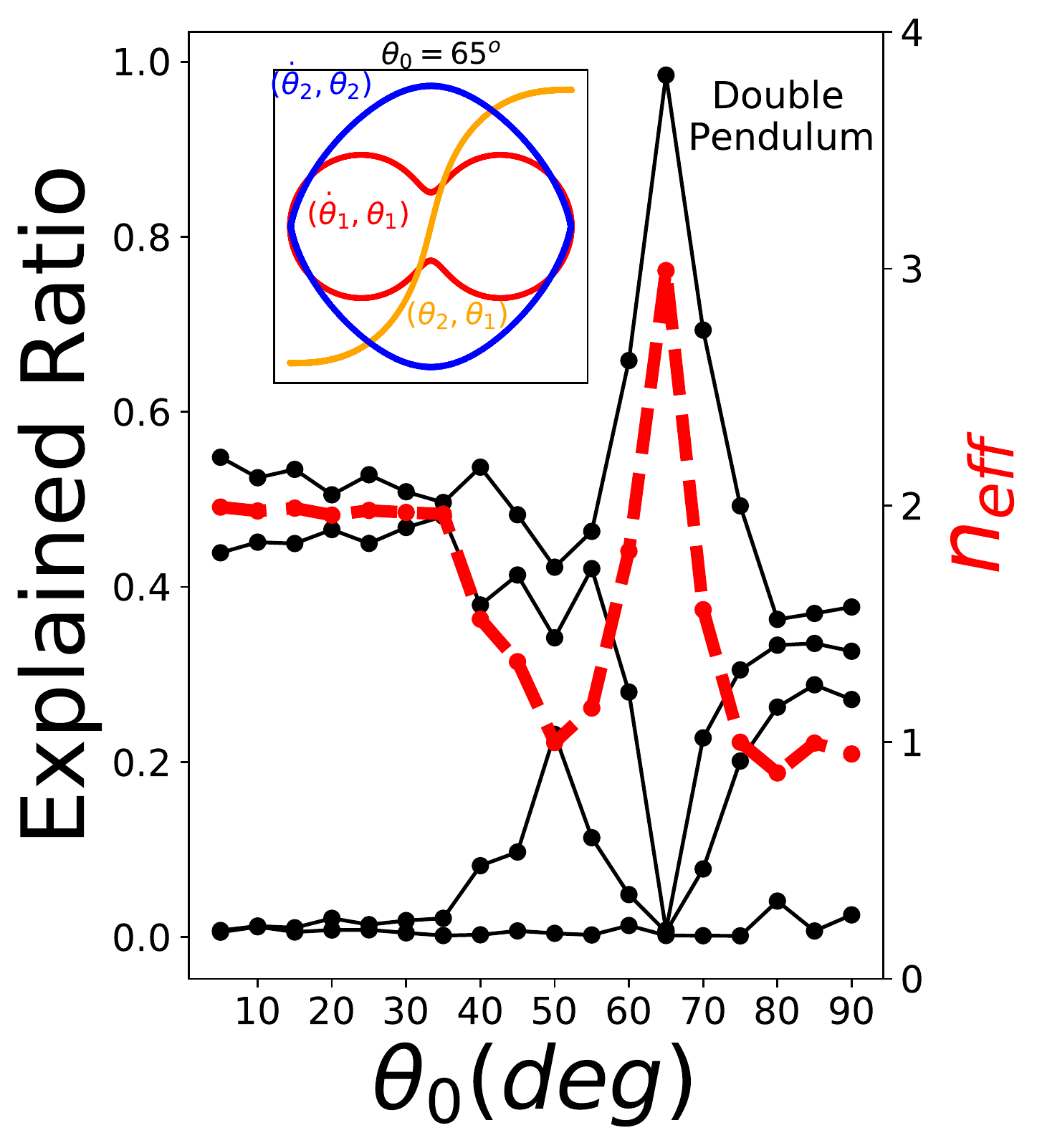}
  	    \vskip-2mm
            \caption{Double Pendulum}
    \end{subfigure}
    \vskip-3mm
    \vskip\baselineskip
    \begin{subfigure}[b]{0.23\textwidth}
            \centering \includegraphics[width=\textwidth]{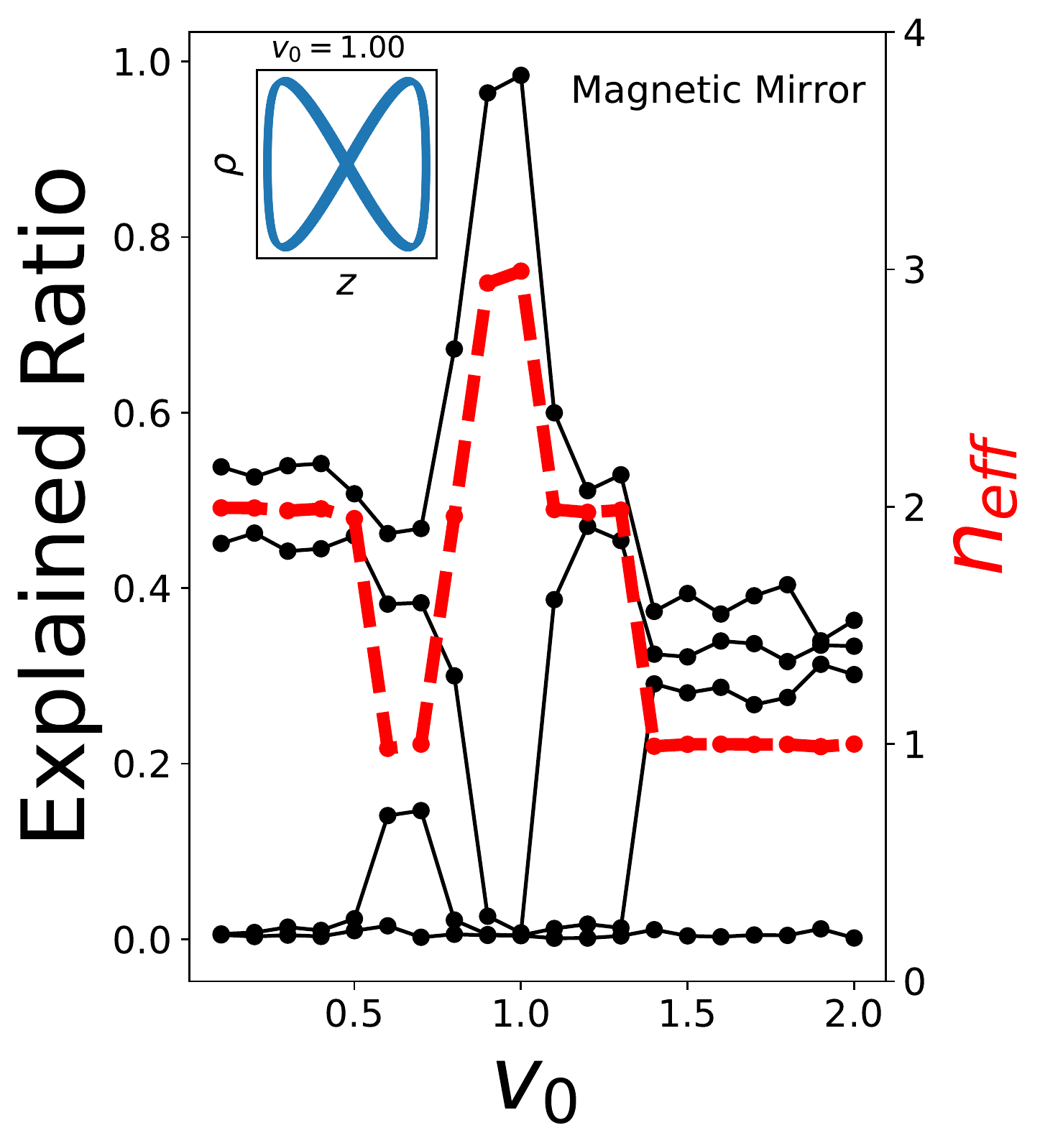}
  	    \vskip-2mm
            \caption{Magnetic Mirror}
    \end{subfigure}
    \hfill
    \begin{subfigure}[b]{0.23\textwidth}  
            \centering 
            \includegraphics[width=\textwidth]{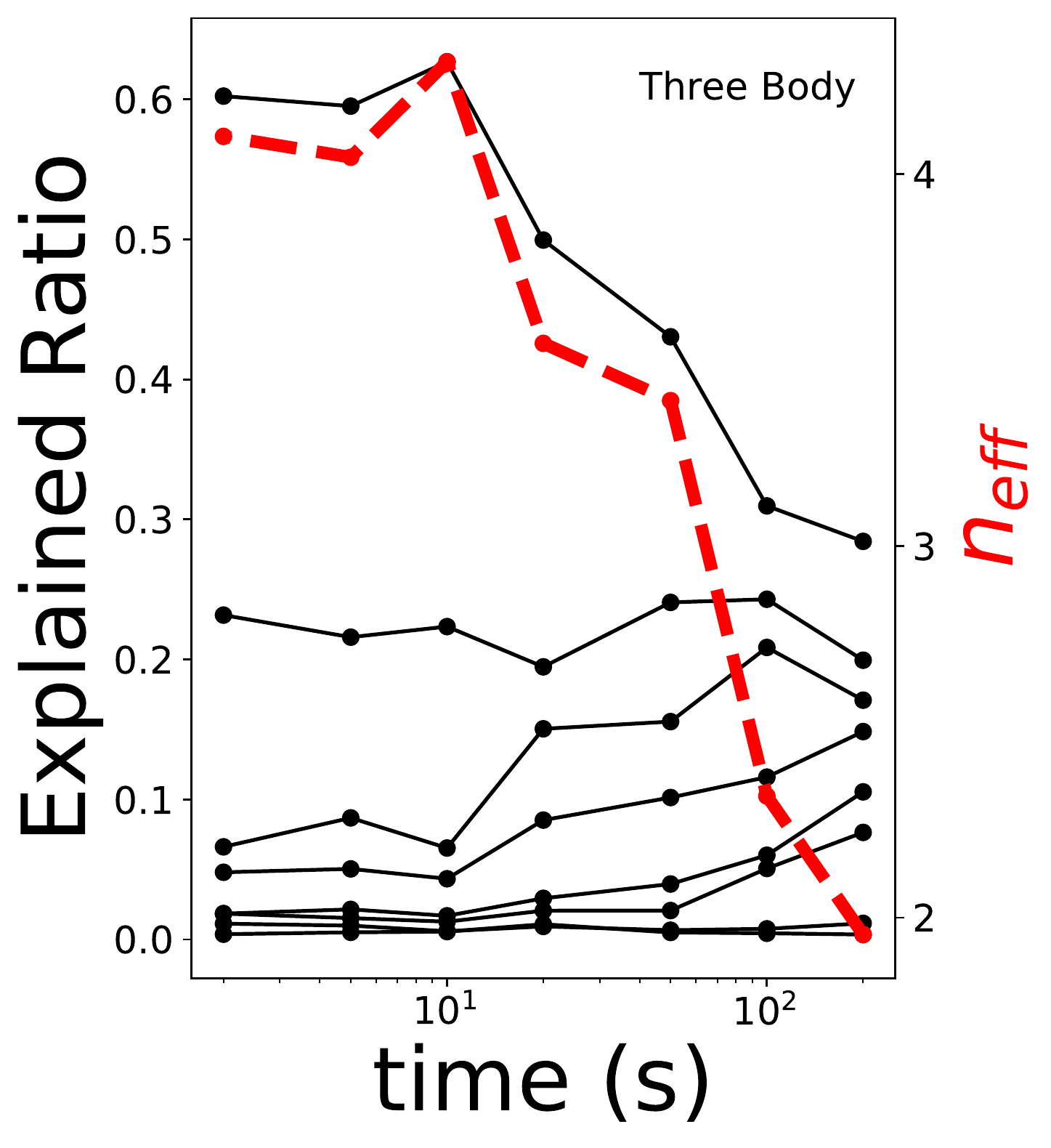}
  	    \vskip-2mm
            \caption{Three-body Problem}
    \end{subfigure}
        
    \caption{AI Poincar\'{e} detection of phase transitions, approximate conserved quantities and periodic orbits, using $\neff$ as an order parameter.}
    \label{fig:phase_transition}
    \vspace{-10pt}
\end{figure}

{\bf Phase transition discovery}
We will now explore how AI Poincar\'{e} can auto-discover not only \textit{exact} conservation laws as above, but also \textit{approximate} ones, revealing physically interesting phase transitions.
There are many reasonable ways of defining an  {\it effective number of conserved quantities} $\neff$ as a smooth function of the explained ratios $w_i$. We want each explained ratio $w_i$ to contribute 1 for small $w_i$ and 0 for $w_i\simgt 1/N$, so here we make the simple choice
\beq{neffEq} 
\neff\equiv\max_L\neff(L),\quad \neff(L)\equiv \sum_{i=1}^N c[\pi N\omega_i(L)],
\eeq
where $c(x)\equiv\cos x$ if $x<\pi/2$, vanishing otherwise. This is seen to agree well with our threshold criterion 
(\fig{fig:hamiltonian}, bottom). Below we fix $L=0.1$ instead of maximizing over it to save computation time.
Let us view $\neff$ as an {\it order parameter} of the dynamical system and consider phase transitions in the parameter space spanned by timescale, initial conditions and Hamiltonian modifications. 

For the {\bf Kepler problem}, we generalize the inverse square force to the form 
$F\propto r^{-(2+\epsilon)}$, $|\epsilon|\ll 1$. This causes the previously conserved Runge-Lenz vector (the major axis direction) to precess by an angle $\sim\epsilon$ per orbit~\cite{wells2011effective}, 
so that its approximate conservation breaks down after $\sim\epsilon^{-1}$ orbits. 
AI Poincar\'{e} is seen to auto-discover this phase transition (\fig{fig:phase_transition}a) 
without using any of the aforementioned physics knowledge.

The {\bf double pendulum} is known to have a regular phase at low energy and a chaotic phase at high energy~\cite{STACHOWIAK2006417}, both with $\neff=1$ (conserving only total energy). 
We change the initial conditions to $\theta_1=\theta_2=\theta_0$ 
and plot the dependence of $\neff$ on $\theta_0$ (\fig{fig:phase_transition}b), which is seen to reveal two additional phases.
(1) An interesting periodic orbit  ($\neff=3$) is discovered at $\theta_0\approx 65^\circ$ (see inset figure), by adjusting initial conditions to maximize $\neff$.
(2) For small $\theta_0$, the small angle approximation allows the system to be accurately linearized and decoupled into two non-interacting normal modes, whose energies are separately conserved ($\neff=2$).

The {\bf Magnetic mirror} is also known to have a regular low-energy phase and a chaotic high-energy phase~\cite{Contopoulos2016AnalyticalSO}, both with $\neff=1$ (conserving only energy). 
We change the initial conditions to  $\dot{\rho}=\dot{z}=v_0/\sqrt{2}$
and the dependence of $\neff$ on $v_0$ in \fig{fig:phase_transition}c is seen to reveal two additional phases.
(1) An interesting periodic orbit  ($\neff=3$) is discovered at $v_0\approx 1.0$ (see inset figure).
(2) At low energy (small $v_0$), the magnetic moment is an adiabatic invariant, so $\neff=2$.

For the three-body problem, we consider initial conditions akin to a tight binary star pair orbiting a more distant star. \fig{fig:phase_transition}d reveals that
the local energy and angular momentum of the tight binary are approximately conserved initially, increasing $\neff$ by 2, until tidal interactions with the distant star eventually cause this conservation to break down.

\section{Conclusions}\label{sec:conclusions}

We have presented {\it AI Poincar\'{e}}, a machine learning algorithm for auto-discovering conserved quantities using trajectory data from unknown dynamical systems.
Tests on five Hamiltonian systems showed that it auto-discovered not only the number of conserved quantities, but also periodic orbits, phase transitions and breakdown timescales for approximate conservation laws.
AI Poincar\'{e} is universal in the sense that it does not require any domain knowledge or even a physical model of how the trajectories were produced. 
It may therefore be interesting to apply it to raw experimental data, for example measured neuron voltages in {\it C.~elegans}. Another promising future direction is improved discovery of symbolic formulas for the discovered conserved quantities by transferring learned geometric information from AI Poincar\'{e} to AI Feynman, \eg, by requiring that symbolic gradient directions match those of our learned pull function.

{\bf Acknowledgements} We thank Qihao Cheng, Artan Sheshmani and Huichao Song  for helpful discussions and the Center for Brains, Minds, and Machines (CBMM) for hospitality. This work was supported by The Casey and Family Foundation, the Foundational Questions Institute, the Rothberg Family Fund for Cognitive Science and IAIFI through NSF grant PHY-2019786.

\bigskip
\bigskip
\bigskip
{\LARGE\bf  Supplementary material}
\appendix

\setcounter{secnumdepth}{2}
 
\section{Preprocessing details}
\label{app:preprocess}

As mentioned in the main text, our AI Poincar\'e method begins with a pre-processing step consisting of prewhitening and optional dimensionality reduction.
The prewhitening performs an affine transformation such that the points in $\T$ obtain zero mean and covariance matrix $\I$, the identity matrix. 
If any eigenvalues of the original covariance matrix vanish, then the corresponding eigenvectors $\e_i$ define linear conserved
quantities $H_i(\x)=\e_i\cdot\x$, and we remove these dimensions before proceeding. 
In practice, we consider an eigenvalue $\lambda$ to vanish if 
\beq{vanishingEigenvalueEq}
|\lambda_i|<\epsilon_p \max_i \lambda_i,
\eeq
where $\epsilon_p=0.001$ is the only tunable hyperparameter in AI Poincar\'e aside from those associated with neural network architecture and training. The absolute value is needed in \eq{vanishingEigenvalueEq} because, although a  covariance matrix is by definition positive semidefinite, numerical errors in the eigenvalue computation can produce small negative numbers. 
Below, we provide motivation and further details for the prewhitening and dimensionality reduction.

\subsection{Why prewhitening helps}

As mentioned in the main text, the dimensionality of the manifold $\M$ can be read off from the explained-ratio diagram at an intermediate range of length scales $L$ that is sandwiched between two phase transitions at $L_a\ll 1$ and $L_b\sim 1$. 
In the other two phases, all explained ratios are $\sim 1/n$ because the Monte Carlo sample covariance matrix $\sim\I$: on large scales because almost the whole (prewhitened) manifold is sampled and on small scales because of roughly isotropic noise. 
Thus $L_a$ is determined by the noise lengthscale, and $L_b$ is determined roughly by the curvature radii of the manifold.  If a manifold is highly anisotropic, it can have quite different curvature radii in different directions, replacing a sharp easy-to-spot transition at $L_b$ by a very slowly rising curve. This suggests that prewhitening can help by making the manifold more isotropic, thus pushing its the different curvature scales closer together, far from the noise scale, 
widening the interval $[L_a,L_b]$ from which we try to measure the manifold's dimensionality. 

\begin{figure}[htbp]
	\centering
	\includegraphics[width=0.8\linewidth]{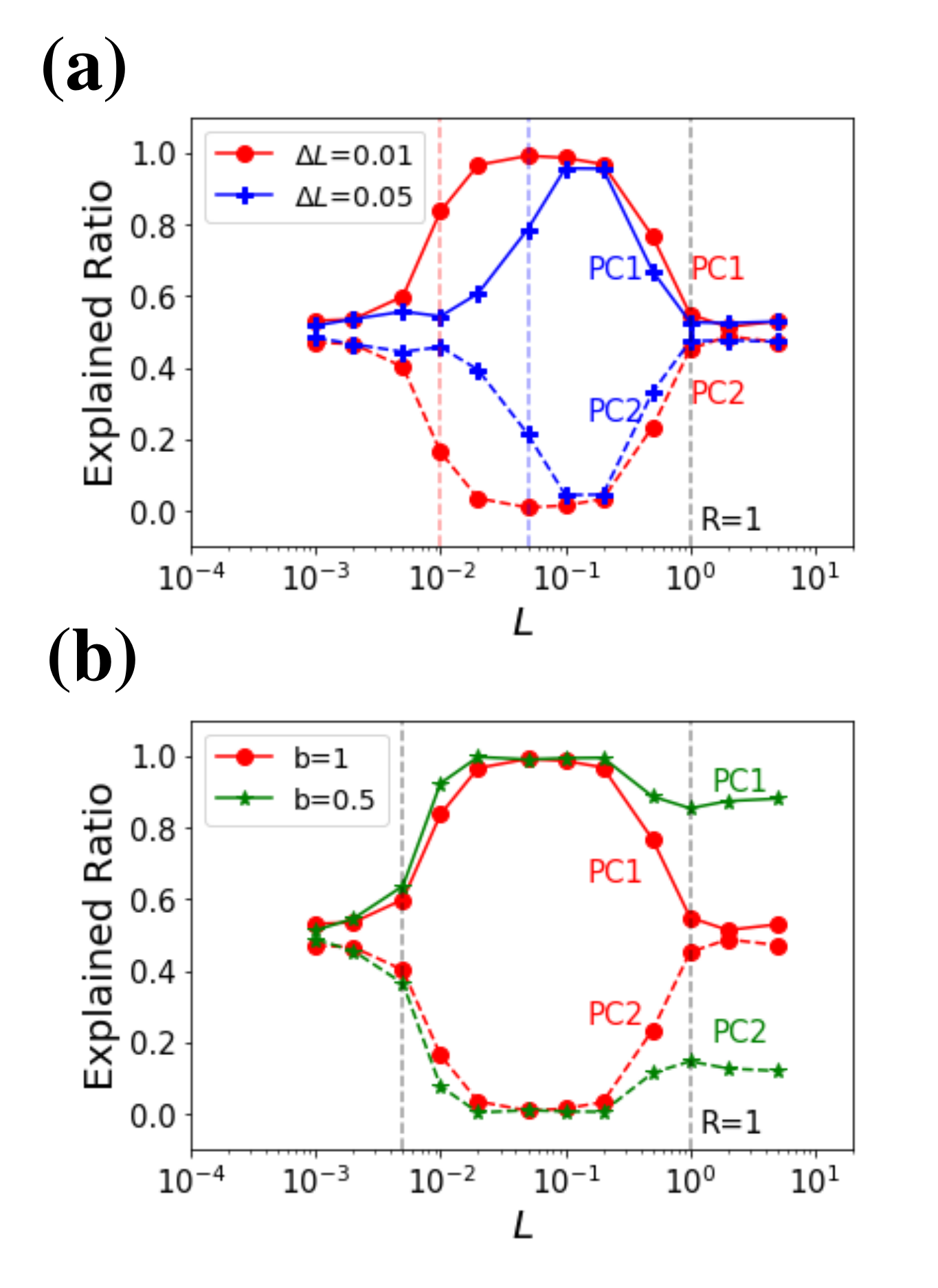}
	\vskip-3mm
	\caption{AI Poincar\'{e} applied to ellipse with noise scale $\Delta L$ and pricipal axes $1$ and $b$, showing that 
(a) the location of the first phase transition is determined by the noise scale $\Delta L$ and (b) the sharpness of the second phase transition is determined by the axis ratio of the ellipse, motivating our prewhitening procedure.}
\label{fig:ai_poincare2}
\end{figure}

This intuition is supported by the numerical experiment results shown in \fig{fig:ai_poincare2}.
Here we test AI Poincar\'{e} without the pre-processing module, using a toy dataset where the manifold is an ellipse with semimajor axis $a=1$ and semiminor axis $b$, \ie, with eccentricity $e = \sqrt{1-b^2}$.
To this, we add Gaussian noise with standard deviation $\Delta\L$.
The top panel shows that $L_a\sim\Delta L$ as expected, \ie, that the leftmost phase transition occurs at the noise scale.
The bottom panel shows why prewhitening helps: when the ellipse is more anisotropic, the second phase transition becomes less prominent and therefore harder to detect. In the limit where $b\to 0$, the ellipse would collapse toward a line and thus contribute to one rather than two principal components. 

\subsection{Robustness to dimensionality reduction and noise} 

\begin{figure}[htbp]
	\includegraphics[width=0.8\linewidth]{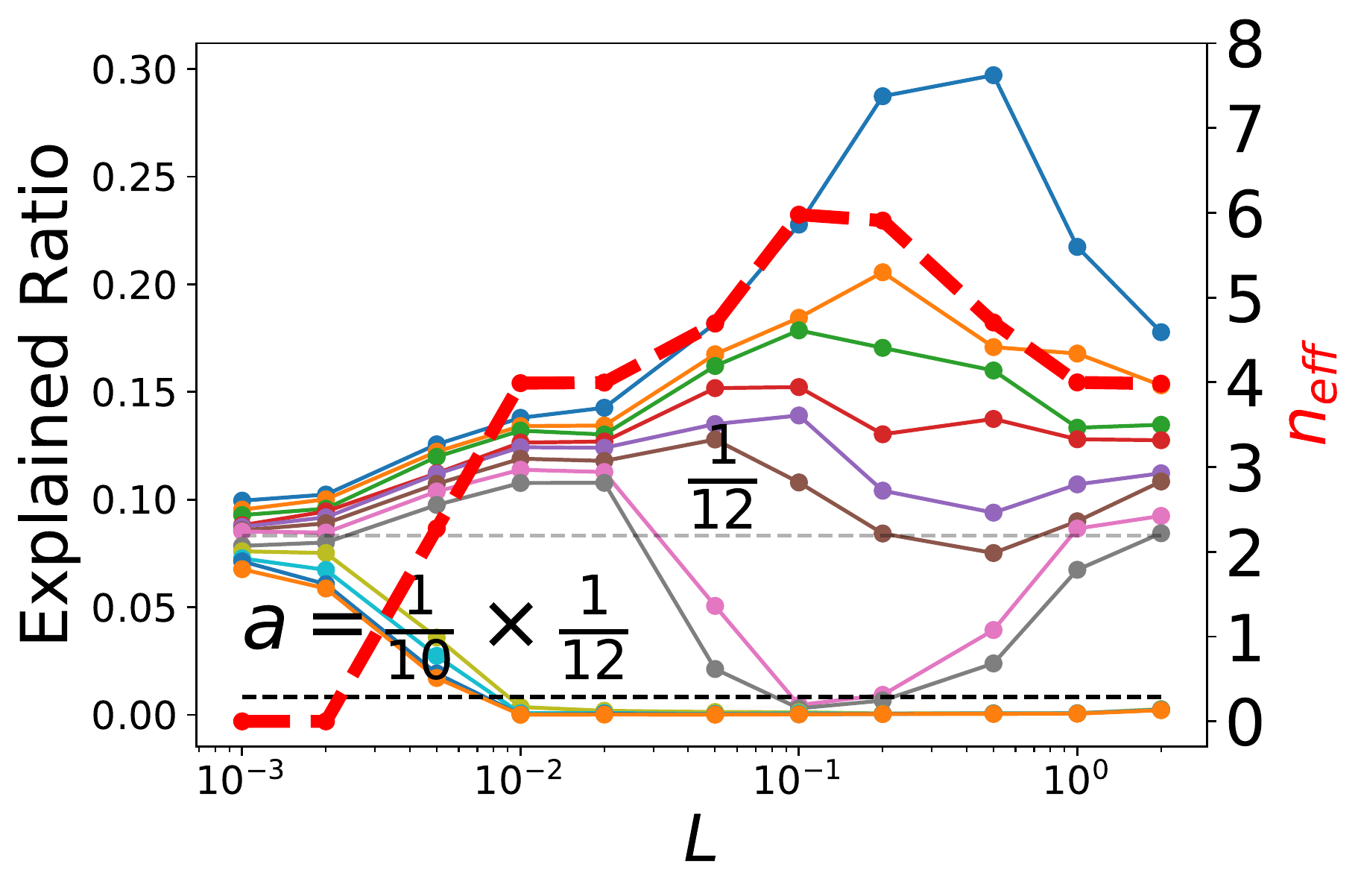}
	\caption{Explained ratio diagram for the three body problem when linear conserved quantities are not removed in preprocessing (see text), showing that AI Poincar{\'e} nonetheless discovers all $n=6$ conserved quantities at $L=0.1, 0.2$ and all $n=4$ linear conserved quantities even at $L=1,2$.}
	\label{fig:threebody_nopca}
\end{figure}
In this subsection, we demonstrate that although dimensionality reduction is useful and simplifies subsequent calculations, our basic AI Poincar\'{e} method is robust enough that it can work even without it.
\Fig{fig:threebody_nopca} shows the three-body problem reanalyzed without removing the four principal components corresponding to linear
conserved quantities, correctly discovering all $\neff\approx 6$ conserved quantities at $L\in[0.1,0.2]$ and asymptoting to $\neff\approx 4$ at $L\simgt 1$, since four linear conserved quantities survive even at large scales.
To avoid division by near-zero in the prewhitening process, we modified it for this experiment so that each principal component was divided not by its standard deviation $\lambda_i^{1/2}$ but by $\lambda_i^{1/2}+\epsilon_n$, where $\epsilon_n=0.001$.

\begin{figure}[htbp]
	\includegraphics[width=0.8\linewidth]{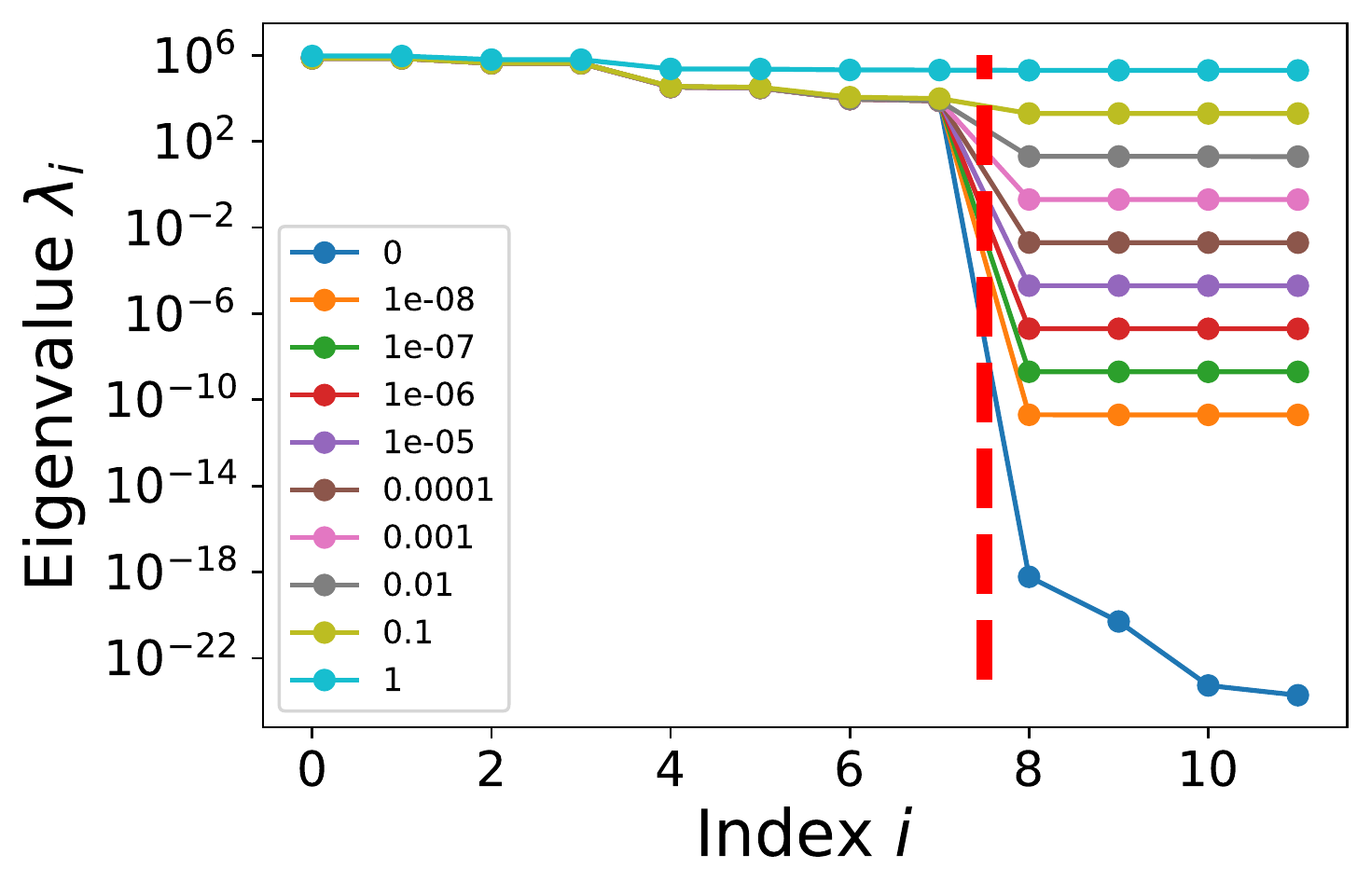}
	\caption{Three body problem: noise dependence of PCA eigenvalues. }
	\label{fig:threebody_preprocessing}
\end{figure}

Let us also briefly comment on the robustness of our method to noise.
\Fig{fig:threebody_preprocessing} illustrates the effect of adding independent random noise to all data points before the prewhitening step.
As long as the added to all data points is drawn from a distribution with the same variance $\sigma^2$, the noise covariance matrix $\sigma^2\I$ commutes with the data covariance matrix, so the effect of the noise is simply to add $\sigma^2$ to all eigenvalues.
We see that this 3-body problem example allows the four linearly conserved quantities (corresponding to the four smallest eigenvalues) to be easily recovered as long as the noise level $\sigma\simlt 0.01$.

\section{Monte Carlo Module}\label{app:monte carlo}

\begin{figure}[htbp]
	\centering
	\includegraphics[width=0.8\linewidth]{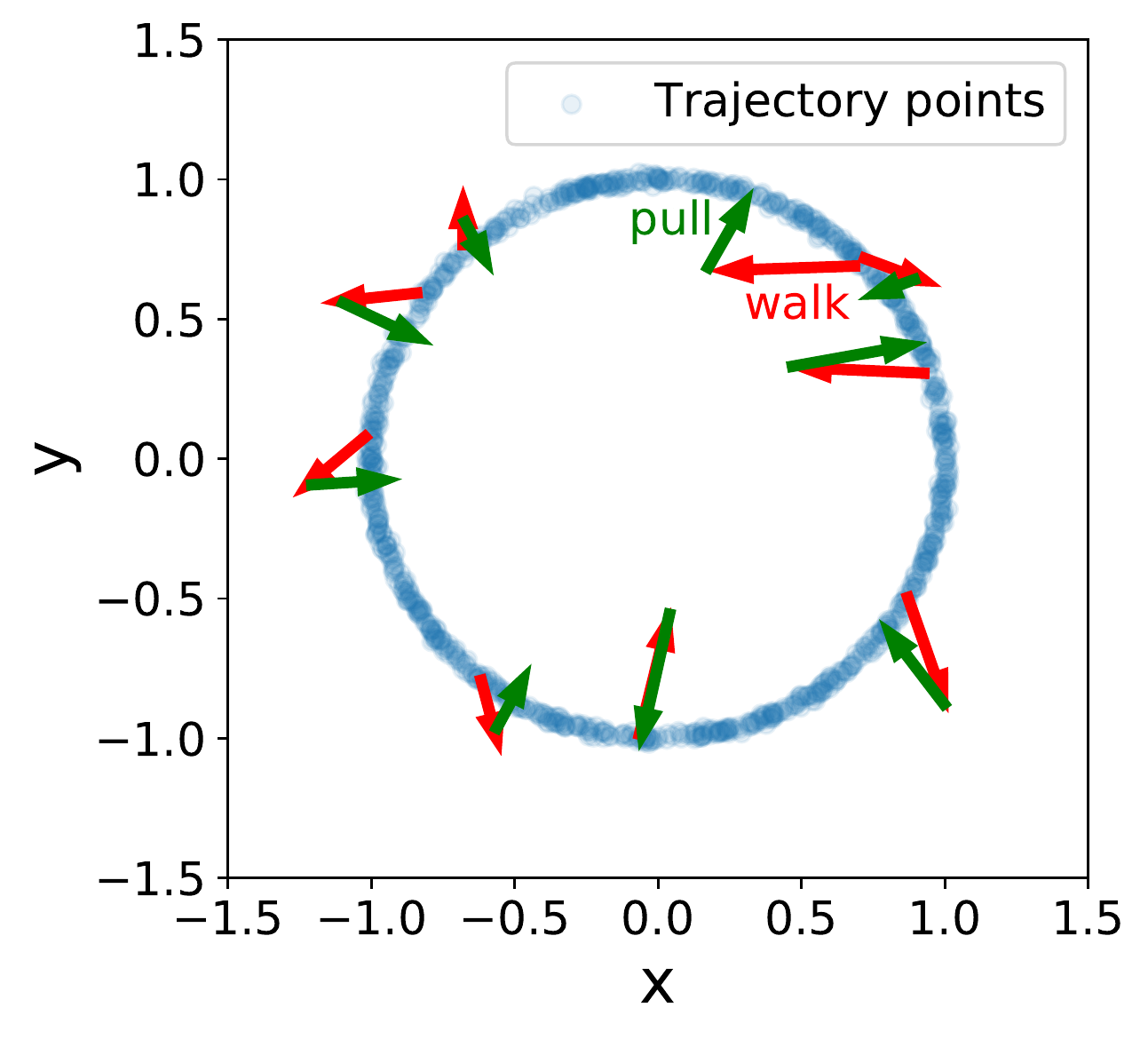}
	\caption{Monte Carlo module illustration: walk and pull steps on a circle. The walk step randomly perturbs points off the manifold (the circle), after which the pull network projects them back toward the manifold.}
	\label{fig:circle_illustrate}
\end{figure}

In this section, we provide additional technical details about our implementation of AI Poincar\'e's Monte Carlo module.
Our method takes input $N_s$ points on a trajectory, which is split into two by using the odd/even points for training/testing, respectively, and used to train the walk/pull network as described in the main text. 
Consequently, the pull network learns about the manifold in a \textit{global} sense, from the full trajectory.
\Fig{fig:circle_illustrate} illustrates this walk/pull step on a circle. Intuitively, the pull network learns to pull points to the closest point on the manifold, which makes the combined walk and pull step approximate a random walk on the manifold.

\begin{figure*}[htbp]
	\includegraphics[width=0.9\linewidth]{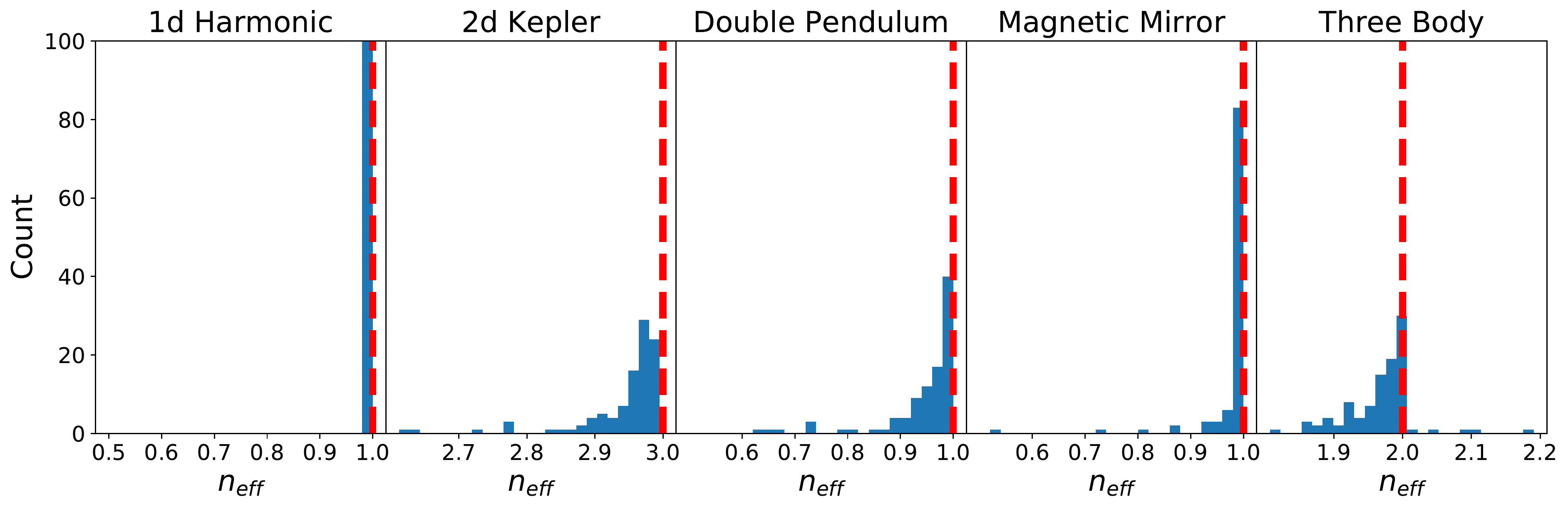}
	\caption{Stability tests of AI Poincar{\'e} for the results in FIG.~\ref{fig:hamiltonian}. The predictions by AI Poincar{\'e} (blue) have relatively small deviations from ground truth (dashed red line), so rounding to the nearest integer provides excellent error correction.
}
\label{StartingPointFig}
\end{figure*}
At inference time (after training this pull network), we randomly select {\it one} starting point (by default the midpoint of the trajectory) and perform $10^3$ walk/pull steps to obtain samples, the principal components of which are \textit{local} in the sense that they depend on the starting point. 
We do this merely to save computer time, since the starting point can be any point on the manifold, and $\neff$ can be averaged for many starting points. 

To test the stability of AI Poincar{\'e} against different choices of starting point,  we randomly select 100 points on the trajectory for testing. At each point $\x$ and at each noise scale $L$, we implement 1,000 walk and pull steps to obtain 1,000 samples, and then apply PCA to obtain explained ratios and calcuate $\neff(\x,L)$. For each point $\x$, we select the $L$ that maximizes $\neff$, \ie, we define $\neff(\x)\equiv\underset{L}{\mathrm{max}}\ \neff(\x,L)$. 
\Fig{StartingPointFig} shows the histogram of $\neff(\x)$ for the 100 random starting points, revealing that the standard deviation of $\neff$ is quite small. 
If we round $\neff(\x)$ to the closest integer, we obtain the correct number of conserved quantities
100 out of 100 times for all five of our physical systems. 

\section{Method comparisons}

\label{sec:baseline}
\begin{figure*}
	\includegraphics[width=0.9\linewidth]{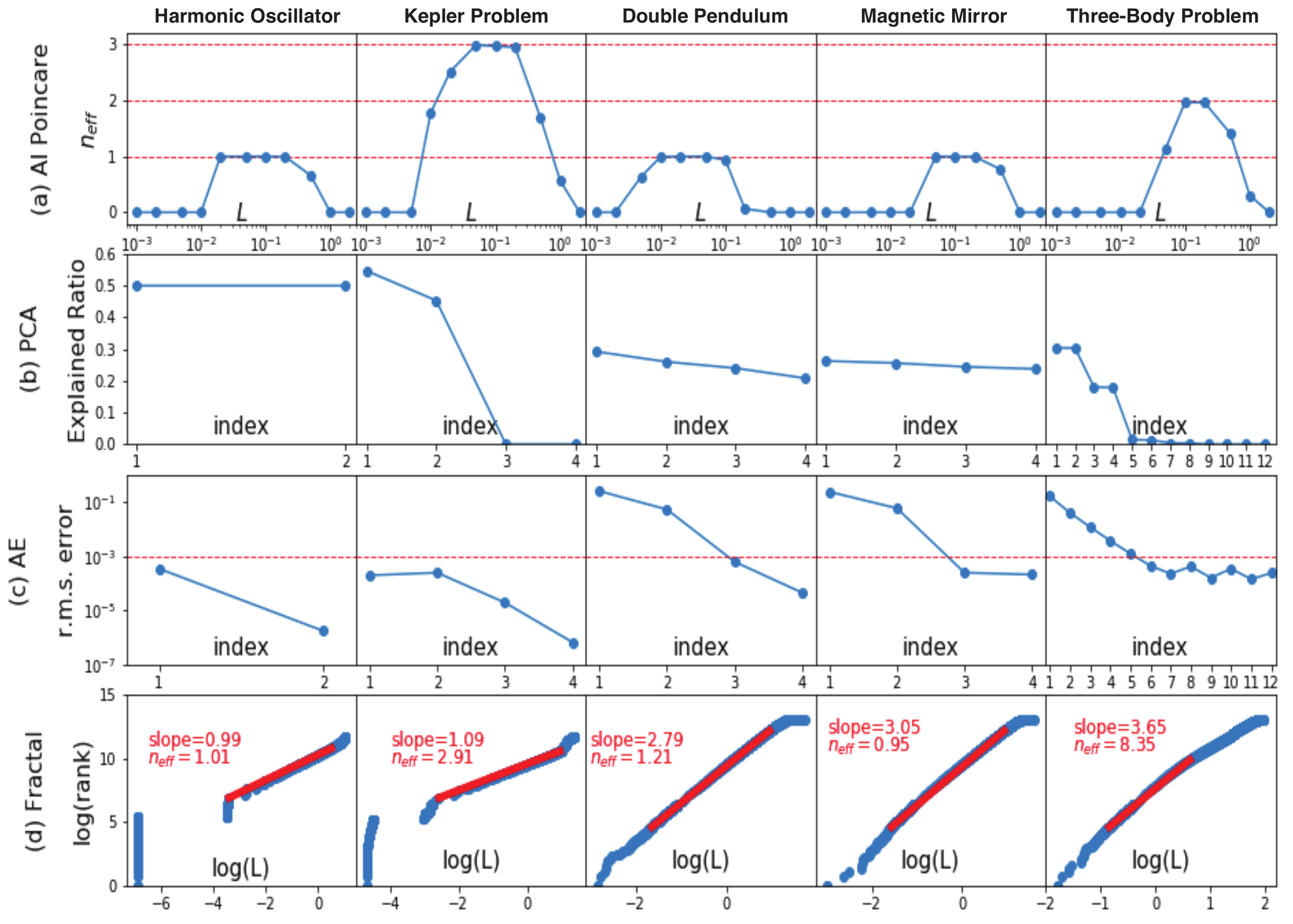}
	\caption{Performance of the (a) AI Poincar\'{e}, (b) PCA, (c) Auto-encoder (c) Auto-encoder and (d) Fractal methods for conserved quantify discovery.
The ground truth $\neff$ is 1 (Harmonic Oscillator), 3 (Kepler Problem), 1 (Double Pendulum), 1(Magnetic Mirror) and 6 (Three Body Problem; 2 after linear dimensionality reduction).}
	\label{fig:baseline}
\end{figure*}

In this section, we compare the performance of AI Poincar\'{e} with the following three other methods for estimating the dimensionality 
$s$ of a state-space manifold $\M$ by analyzing the point set $\T\subset\M$.
\begin{enumerate}
\item {\bf PCA}: A popular method for dimensionality estimation is to perform a principal component analysis of the covariance matrix of the points and estimate the dimensionality as the number of large eigenvalues.
\item {\bf Auto-encoder}: An auto-encoder is simply a non-linear generalization of PCA, where the input vector $\x\in\mathbb{R}^N$ is transformed nonlinearly (but smoothly) to $\y\in\mathbb{R}^s$ and then transformed back to $\x'\in\mathbb{R}^N$ in such a way that average reconstruction error $|\x'-\x|^2$ is minimized. The dimensionality can then be estimated as the the smallest $s$ that results in accurate reconstruction. 
\item {\bf Fractal Method}: If a manifold has inherent dimensionality $s$, one would expect the number of point pairs separated by less than $L$ to scale as $N\propto L^s$  for small $L$, so the dimensionality (even if fractal, \ie, non-integer) can be estimated as the slope of  $\log N$ as a function of $\log L$.
\end{enumerate}
\Fig{fig:baseline} shows the results of applying AI Poincar\'{e} and these three methods to each of our five physics examples.
Unsurprisingly, PCA performs the worst: because it is simply a {\it linear} autoencoder, it can only discover linear conserved quantities (that confine the data to a hyperplane), and therefore discovers only the four linearly conserved quantities for the 3-body problem.
A trivial example where PCA fails is a circle: it has two equally large principal components even though it is a one-dimensional manifold.
Note that, in contrast, AI Poincar\'{e} uses PCA for dimensionality estimation {\it locally}, not globally as we have done here.

For the fractal method, we manually identify an intermediate range of scales where the slope 
$d\log N/d\log L$ is roughly constant and estimate dimensionality from that. We do this because, 
as described in the main text, we expect the effective dimensionality to depend on scale, approaching $N$ on small scales (because of noise) and large scales (because of manifold curvature).  \Fig{fig:baseline} shows that, after some manual tuning of these ranges, we were able to get the correct dimensionality for four of the five cases. The fractal method is slower than AI Poincar\'{e} because the number of 
pairs scales as the square of the number of points, so we used a random subset $10^3$ random points for this figure.

For the auto-encoder, \fig{fig:baseline} shows that we are able to obtain the correct manifold dimensionalities (and hence the correct number of conserved quantities) if we manually tune the threshold defining ``accurate" as $10^{-3}$. Unfortunately, the figure also shows that
the inferred dimensionalities depend strongly on the choice of this threshold, and that this choice is far from obvious in cases such as that of the double pendulum: the reconstruction error curves lack sharp and clear-cut phase transitions that can guide this threshold choice. 
This may be at least in part due to topological problems, since topologically non-trivial manifolds may require extra dimensions to auto-encode with a continuous mapping.
In contrast, every single one of the five AI Poincar\'{e} $\neff$ curves is seen to have an obvious flat plateau, so the simple criterion
$\neff\equiv\max_L\neff(L)$ works every time.
Another advantage of  AI Poincar\'{e} over the other methods is that it can measure dimensionality also {\it locally}, at each part of the manifold separately.

\begin{table}[ht]
	\centering
	\caption{Training Loss/Testing Loss}
	\begin{tabular}{|l|c|c|c|}\hline
		Model & $L=0.01$ & $L=0.1$ & $L=1$ \\\hline
		Harmonic   & $.000073/.000072$ & $.0033/.0038$ & $0.42/0.40$ \\
		Kepler & $.000059/.000057$ & $.00074/.00083$ & $0.41/0.39$ \\
		Double Pendulum  & $.000099/.000097$ & $.0076/.0071$ & $0.44/0.44$ \\
		Magnetic Mirror & $.00012/.00011$ & $.0068/.0066$ & $0.48/0.51$ \\
		Three Body 
		& $.000099/.00011$ & $.0068/.0065$ & $0.46/0.42$\\\hline
	\end{tabular}
	\label{tab:poincare_no_overfit}
\end{table}

\section{Does AI Poincar\'e overfit?}

To test whether AI Poincar\'{e} has a tendency to overfit, we train the pull network using only odd-numbered  trajectory points, while holding back the even-numbered points for testing, and
tabulated the AI Poincar\'{e} training loss and testing loss for all five of our physical systems at various length scales. As can be seen in Table~\ref{tab:poincare_no_overfit}, this reveals no evidence of overfitting where testing loss is systematically larger than training loss.

\section{Does AI Feynman overfit?}

As described in the main text,  AI Feynman is trained to discover a formula that is as constant as possible (=1 and 2, respectively) on two different trajectories. To test for overfitting, 
we quantify how how constant the discovered formula is by computing its mean and standard deviation)along a third trajectory not seen at training time.
The results in Table~\ref{tab:feynman_no_overfit} indicate no evidence of overfitting, \ie, that the test error is systematically larger than the training error. 

Our criterion for whether AI Feynman succeeds in finding the correct formula is that 
the discovered formula is mathematically identical to the ground truth formula up to 3\% variations in fitted numerical coefficients.
For example, the ground truth is that the harmonic oscillator has energy $H=\frac{1}{2}(x^2+p^2)$. Since $H'\equiv aH+b=\frac{1}{2}ax^2+\frac{1}{2}ap^2+b$ is also conserved, we consider the discovered equation as correct if $H'$ has the functional form $H'=cx^2+dp^2+e$ and $|{c\over d}-1|<3\%$.

\begin{table*}[bht]
	\centering
	\caption{How empirically conserved are the symbolic conserved quantities found by AI Feynman?}
	\begin{tabular}{|l|c|c|c|c|}\hline
	Conserved quantity & Formula found?& Trajectory 1 (training) &  Trajectory 2 (training) & Trajectory 3 (testing) \\\hline
	Harmonic: energy & Yes & $1.002\pm 0.005$ & $2.001\pm 0.010$ & $1.457\pm 0.003$ \\
	Kepler: energy & Yes & $1.02\pm0.03$ & $1.97\pm0.05$ & $1.35\pm0.04$ \\
	Kepler: angular momentum & Yes & $1.01\pm0.01$ & $1.98\pm0.03$ & $1.48\pm0.02$ \\
	Kepler: Runge-Lenz vector & No & & & \\
        Double pendulum ($|\theta_0|\ll1$): energy & Yes & $1.04\pm 0.07$ & $1.97\pm 0.05$ & $1.48\pm0.06$\\
	Double pendulum ($|\theta_0|\sim 1$): energy & No & & & \\
	Magnetic mirror: energy & Yes & $1.02\pm0.04$ & $2.03\pm0.05$ & $1.45\pm0.07$ \\
	Three-body: energy & No & & & \\
	Three-body: angular momentum & Yes & $1\pm0.02$ & $2\pm 0.03$ & $1.77\pm0.02$ \\
	Three body: $x_c$ & Yes & $1\pm .00002$ & $2\pm .00005$ & $1.354\pm .00004$ \\
	Three body: $y_c$ & Yes & $1\pm .00004$ & $2\pm .00003$ & $1.445\pm .00003$ \\
	Three body: ${\dot x}_c$ & Yes & $1\pm .00003$ & $2\pm .00002$ & $1.327\pm .00002$ \\
	Three body: ${\dot x}_y$ & Yes & $1\pm .00008$ & $2\pm .00002$ & $1.813\pm .00004$\\
	\hline
	\end{tabular}
	\label{tab:feynman_no_overfit}
\end{table*}

\bibliography{poincare}

\end{document}